\pdfoutput=1

\documentclass[11pt]{article}

\usepackage[]{EMNLP2023}

\usepackage{times}
\usepackage{latexsym}

\usepackage[T1]{fontenc}

\usepackage[utf8]{inputenc}

\usepackage{microtype}

\usepackage{enumitem}
\usepackage{bookmark}
\usepackage{bm}
\usepackage{amsfonts}
\usepackage{subfigure}
\usepackage{graphicx}
\usepackage{booktabs}
\usepackage{makecell}
\usepackage{amsmath}
\usepackage{arydshln}
\usepackage{multirow}
\usepackage{stmaryrd}
\usepackage{color}
\usepackage{siunitx} 

\usepackage{inconsolata}

\title{Benchmarking Large Language Models with Augmented Instructions for Fine-grained Information Extraction}

\author{ Jun Gao\textsuperscript{1}\thanks{\;\;Work done when Jun Gao was interning at 4Paradigm.} \enskip Huan Zhao\textsuperscript{2} \enskip Yice Zhang\textsuperscript{1} \enskip Wei Wang\textsuperscript{3} \enskip Changlong Yu\textsuperscript{4} \enskip Ruifeng Xu\textsuperscript{1}\\
 \textsuperscript{1}Harbin Institute of Technology (Shenzhen) \quad \textsuperscript{2}4Paradigm. Inc.\\
   \textsuperscript{3}Tsinghua University\quad \textsuperscript{4}HKUST, Hong Kong, China  \\
\normalsize \texttt{imgaojun@gmail.com}
}

\begin{document}
\maketitle
\begin{abstract}
  Information Extraction (IE) is an essential task in Natural Language Processing. Traditional methods have relied on coarse-grained extraction with simple instructions. However, with the emergence of Large Language Models (LLMs), there is a need to adapt IE techniques to leverage the capabilities of these models.
  This paper introduces a fine-grained IE benchmark dataset tailored for LLMs, employing augmented instructions for each information type, which includes task descriptions, extraction rules, output formats, and examples. Through extensive evaluations, we observe that encoder-decoder models, particularly T5 and FLAN-T5, perform well in generalizing to unseen information types, while ChatGPT exhibits greater adaptability to new task forms. Our results also indicate that performance is not solely dictated by model scale, and highlight the significance of architecture, data diversity, and learning techniques. This work paves the way for a more refined and versatile utilization of LLMs in Information Extraction.
\end{abstract}

\section{Introduction}

In the field of Natural Language Processing (NLP), Information Extraction (IE) is a pivotal task that aims to identify and extract valuable information from unstructured text. This task encompasses several sub-tasks, including entity extraction~\citep{Yan2021AUG, Wang2021ImprovingNE} and event extraction~\citep{Lin2020AJN}, which play a crucial role in industries such as finance, healthcare, and law, facilitating machines in processing large-scale text.

\begin{figure}[ht] 
  \centering
  \includegraphics[width=0.9\linewidth]{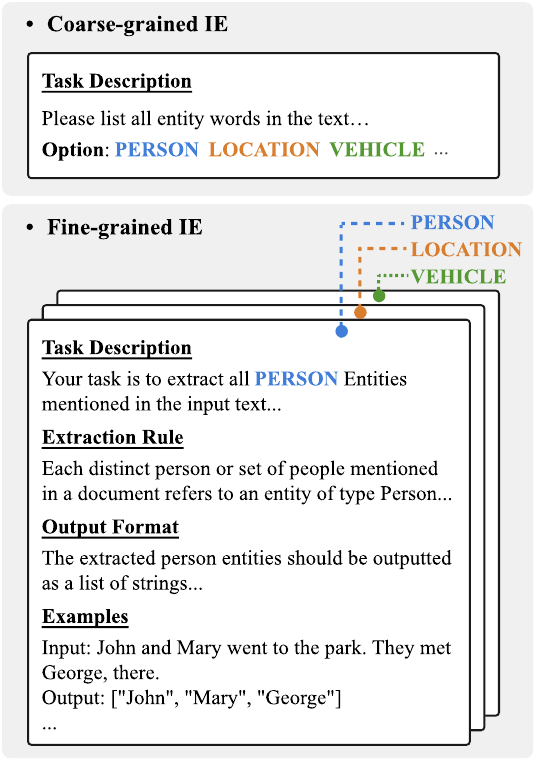}
  \caption{A comparison of the traditional Coarse-Grained IE Instruction~\citep{Lu2022UnifiedSG,wang2023instructuie} with our proposed Fine-Grained IE Instruction, using entity extraction as an illustrative example.}
  \label{fig:example}
\end{figure}

Traditional IE methods predominantly depend on supervised learning~\citep{Lin2020AJN,Du2020EventEB,Lu2021Text2EventCS}, which requires vast labeled datasets for model training. The labeling process can be both time-consuming and expensive, thus creating barriers to the adoption of IE technologies.
In contrast, the advent of Large Language Models (LLMs) such as GPT-3~\citep{Brown2020LanguageMA} has provided an alternative approach. These LLMs demonstrate promising in-context learning capabilities, which could potentially alleviate the need for substantial labeled datasets. This is a remarkable stride forward, as it represents an opportunity to make IE technologies more accessible and efficient.

Despite the potential benefits of LLMs’ in-context learning in IE, existing literature has limitations in evaluating these models' efficacy in IE. Previous studies tend to focus on evaluating performance across a single structure (i.e., either using decoder-only~\citep{Li2023CodeIELC,gao2023exploring,ma2023large,li2023evaluating} or encoder-encoder~\citep{Lu2022UnifiedSG,Liu2023RexUIEAR,wang2023instructuie} models) and on unseen information types~\citep{Lu2022UnifiedSG,wang2023instructuie}. However, there is a dearth of research addressing the generalization performance on different kinds of extraction tasks, such as moving from event extraction to entity extraction.
Moreover, the current approach~\citep{Lu2022UnifiedSG,Liu2023RexUIEAR,wang2023instructuie,Li2023CodeIELC}  in previous work has involved coarse-grained IE tasks, where a simple instruction without detailed extraction guidelines is used to extract multiple information types. 
This neglects vital aspects such as extraction rules, output format descriptions, and illustrative examples, which are crucial for adapting to different information types and tasks.

In this paper, we address these shortcomings by introducing a fine-grained IE benchmark dataset with augmented instructions. The motivation behind transitioning from coarse-grained to fine-grained IE stems from the observation that incorporating detailed extraction guidelines for each information type within the original instructions would cause the instructions to vastly exceed the input length limitations of the model. Fine-grained IE differs from coarse-grained IE in that it treats each information type as a distinct task. Specifically, instead of using a single instruction to extract multiple information types, fine-grained IE employs augmented instructions for each information type, including task descriptions, extraction rules, output formats, and examples. This is depicted in Figure~\ref{fig:example}, which visually contrasts the traditional Coarse-Grained IE with our proposed Fine-Grained IE employing augmented instructions.

A key objective of this study is to stringently evaluate large language models' capabilities in in-context learning for fine-grained IE tasks. We focus on assessing the models' generalization to novel information types and task forms, utilizing a diverse dataset.
We evaluate an array of models, including both \textbf{encoder-decoder} and \textbf{decoder-only} architectures, enabling a thorough analysis of their impact on performance.
Our evaluation encompasses two critical dimensions of generalization:
(1) \textbf{Generalization Across Unseen Information Types}: Models are trained on the same task form but tested on a different information type.
(2) \textbf{Generalization Across Unseen Task Forms}: Models are trained on a partial form of the task and are tested on an entirely different form of the task.

\textbf{Summary of Insights:} The experiment unveils several key insights into the generalization performance of large language models (LLMs) in IE tasks. Encoder-decoder architectures, notably T5 and FLAN-T5, excel in generalizing to unseen information types due to their prowess in capturing input-output relationships. However, they falter in adapting to novel task forms, highlighting a trade-off between specialization and flexibility. ChatGPT, with its decoder-only architecture and in-context learning, demonstrates remarkable adaptability to unfamiliar task structures. Instruction components play a significant role, where \textsc{Extraction Rule} and \textsc{Demonstration Examples} emerge as critical for guiding LLMs effectively, whereas \textsc{Task Description} and \textsc{Output Format} hold variable importance across models. Additionally, the experiment reveals that performance scaling is non-linear with respect to both training data quantity and model size, emphasizing the importance of data diversity and judicious balancing of model scale.


\section{Evaluating In-context Learning in Information Extraction}
In this work, we aim to rigorously evaluate the ability of large language models to perform in-context Learning for fine-grained IE tasks, with a particular emphasis on assessing their generalization capabilities to unseen information types and task forms.

\paragraph{Generalization Over Unseen Information Types.}
In this scenario, models are presumed to have been trained on a diverse set of information types within a particular task structure. They are subsequently evaluated based on their ability to adapt and perform accurately when confronted with novel information types within the same task structure.
Formally, let us represent the set of information types that the model is exposed to during training as $I = \{i_1, i_2, i_3,..., i_n\}$. When the model is presented with a novel information type $i_u \notin I$, we evaluate its capacity to extrapolate its learned knowledge to this new type. For an input text $X$ that contains instances of the new information type $i_u$, the model’s task is to extract these instances. We represent this as $Y = G(X | i_u)$, where $G$ is the function that the model has learned for IE, and $Y$ is the set of extracted information instances.

\paragraph{Generalization Over Unseen Task Forms.}
While models have traditionally been constrained to tasks that closely resemble the structure they were trained on, the emergence of large language models (LLMs) introduces the potential for more adaptable models capable of understanding and adjusting to new task forms.
To formalize this, let's denote the set of task structures the model is trained on as $T = \{t_1, t_2, t_3,..., t_n\}$. Upon encountering a new task form $t_u \notin T$, we assess the model's capability to apply its existing knowledge base to effectively execute the task defined by $t_u$. For an input text $X$, the model is expected to produce an output $Y$ that aligns with the requirements of the new task form. This can be mathematically represented as $Y = F(X | t_u)$, where $F$ represents the function that the model has internalized to map inputs to outputs across different task structures.

\section{Augmented Instructions for Fine-grained Information Extraction}
To achieve a more comprehensive assessment, our focus is on discerning how well these models understand and apply extraction rules and demonstration examples. Given the importance of fine-grained analysis for unearthing specific strengths and weaknesses of LLMs, our dataset considers each type of information as an independent task, requiring meticulous attention to detail.
Specifically, our dataset encompasses an extensive spectrum of information types, including persons, locations, diverse event types, among others. Each of these information types corresponds to a distinct extraction task, such as extracting names of persons or identifying various events described in a text.

\paragraph{Augmented Instruction Schema.}
What sets our fine-grained instructions apart from prior approaches~\citep{Lu2022UnifiedSG,Li2023CodeIELC} is the inclusion of more granular information for each type of information. Instead of just having a task description and output options, our augmented instruction schema integrates extraction rules, specifies output formats, and provides illustrative examples. These additional components are instrumental in equipping the model with an in-depth understanding of the extraction tasks and in standardizing the output for further processing. The instruction schema is composed of the following elements:
\begin{itemize}[leftmargin=*,noitemsep,nolistsep]
\item \underline{\textsc{Task Description}}: A succinct, overarching summary of the task, articulating the primary objective without delving into particulars.
\item \underline{\textsc{Extraction Rule}}: Comprehensive and unambiguous guidelines, formulated in natural language, that outline the specifics of extracting the requisite information from the input text. 
\item \underline{\textsc{Output Format}}: Defines the structural and organizational requirements for the extracted information, offering a systematic template for the model's output. This facilitates uniformity in the presentation of results, which is essential for efficient handling and use of the extracted data.
\item \underline{\textsc{Demonstration Examples}}: Representative input-output pairs that exemplify the correct application of the extraction rules across varied input texts. These examples serve to resolve any potential ambiguities and provide practical demonstrations to reinforce the model’s understanding of the task.
\end{itemize}

\begin{figure*}[ht] 
  \centering
  \includegraphics[width=0.9\linewidth]{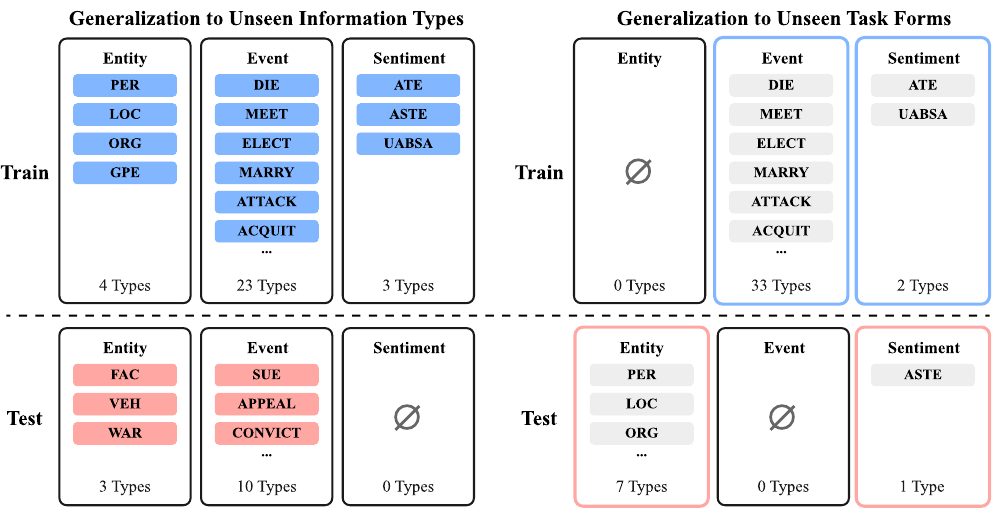}
  \caption{Data division for generalization to Unseen Information Types and Unseen Task Forms. For a detailed view of the data splits, please refer to Figure~\ref{fig:data_split_table} in the Appendix.}
  \label{fig:data_split}
\end{figure*}

\begin{figure}[ht] 
  \centering
  \includegraphics[width=0.9\linewidth]{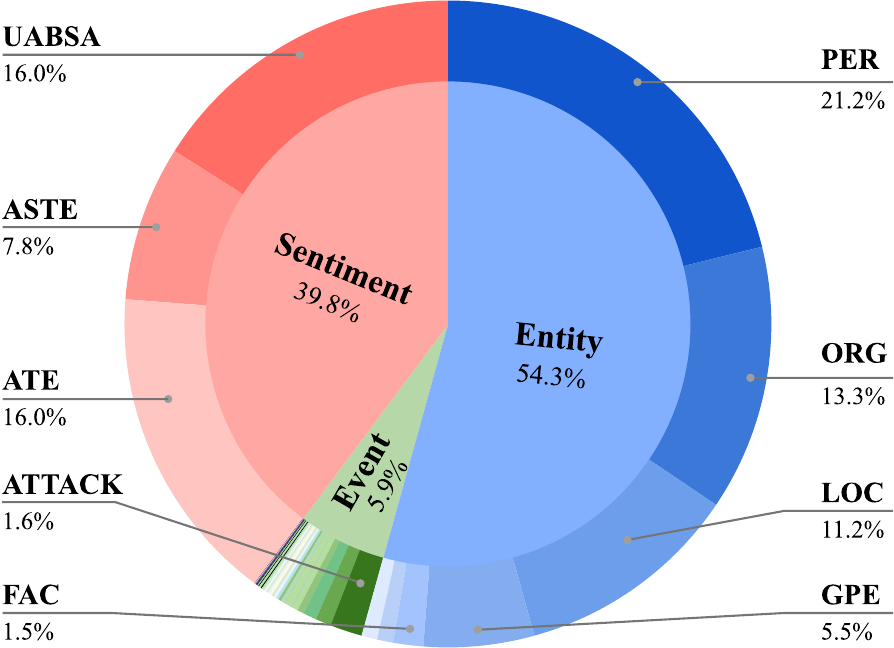}
  \caption{Data Statistics.}
  \label{fig:data_stats}
\end{figure}

\paragraph{Diverse Information Extraction Tasks.}
Building a comprehensive dataset for IE from the ground up can be both resource-intensive and time-consuming. To optimize resource utilization while still achieving a broad coverage, we have amalgamated a selection of pre-existing datasets pertinent to IE. This amalgamation comprises 5 datasets, encompassing three core facets of IE - entity extraction, event extraction, and sentiment analysis. A visual representation of the data distribution is depicted in Figure~\ref{fig:data_stats}. Details of the data construction can be found in Appendix~\ref{sec:data_collection}.

\section{Benchmarking LLMs with Fine-grained IE Tasks}

\subsection{Experimental Setup}
We assess the generalization capabilities of IE models across different facets, namely: Generalization to Unseen Information Types, and Generalization to Unseen Task Forms. Figure~\ref{fig:data_split} shows the dataset partitioning across these dimensions.
\paragraph{Generalization to Unseen Information Types.}
In this scenario, the models are trained on a restricted set of information types and are evaluated on previously unseen information types. The training dataset includes 4 out of 7 entity types, 23 out of 33 event types, and all 3 sentiment information types. For evaluation, we randomly sampled 100 examples for each of the 3 entity types to be tested. Since the number of available samples for each of the event types to be tested was fewer than 100, we utilized the entire dataset for those event types. In total, the test set comprises 700 cases.

\paragraph{Generalization to Unseen Task Forms.}
Here, we evaluate the model's capability to generalize across different forms of IE tasks. Unlike the first setup, where the task form remains the same but the information types differ, here we change the task form itself.
The training set encompasses all event extraction tasks and two of the three sentiment IE tasks, namely ATE and UABSA. The test set includes 100 randomly sampled examples for each of the 7 entity types to be tested and 1,000 randomly sampled examples for the ASTE task, summing up to 1,700 test samples. ASTE extracts aspect, sentiment polarity, and an additional opinion element, making it a higher-order task compared to ATE and UABSA.

For each training sample, we supplement it with 5 randomly sampled examples from the training set, sharing the same type as the demonstration examples. Notably, different training examples are paired with distinct demonstration examples. For the test samples, we include 5 randomly selected demonstration examples in their instructions, ensuring that these demonstration examples are exclusive from the test samples. These demonstration examples remain constant across all test samples. For a detailed view of the data splits, please refer to Figure~\ref{fig:data_split_table} in the Appendix.

\subsection{Models and Evaluation Metrics}
We conducted a comparison between two categories of large language models that are built on different architectures. In the encoder-decoder category, we considered models such as T5~\citep{Raffel2019ExploringTL} and FLAN-T5\citep{chung2022scaling}, both of which are available in sizes of 3 billion (3B) and 11 billion (11B) parameters. In contrast, in the decoder-only category, we looked at models like LLaMa~\citep{touvron2302llama} and BLOOM~\citep{scao2022bloom}, in addition to ChatGPT~\footnote{\url{https://chat.openai.com/}}. LLaMa offers models with 7 billion (7B) and 13 billion (13B) parameters, while BLOOM provides models with 3 billion (3B) and 7.1 billion (7.1B) parameters. Note that the results for ChatGPT were based on testing performed on June 20, 2023. With the exception of ChatGPT, which was able to utilize our instructions directly for in-context learning, the remaining models underwent fine-tuning on our training dataset with fine-grained instructions before being subjected to in-context learning. Implementation details can be found in Appendix~\ref{sec:impl_detail}.

The performance of all models in this task is evaluated using the F1-score as the metric for assessing the accuracy of the information extracted.

\section{Experimental Results}

\subsection{Overall Results}

\begin{table*}[!ht]
  \centering
  \small
  \begin{tabular}{
      ll
      S[table-format=2.2]
      S[table-format=2.2]
      S[table-format=2.2]
      S[table-format=2.2]
      S[table-format=2.2]
      S[table-format=2.2]
  }
  \toprule
      & & \multicolumn{3}{c}{\textbf{Unseen Information Type}} & \multicolumn{2}{c}{\textbf{Unseen Task Form}} \\
      \cmidrule(lr){3-5} \cmidrule(lr){6-7}
      \multirow{-2.5}{*}{\textbf{Structure}} & \multirow{-2.5}{*}{\textbf{Model}} & {\textbf{Entity}} & {\textbf{Trigger}} & {\textbf{Argument}} & {\textbf{Entity}} & {\textbf{ASTE}} & {\textbf{AVG}}\\
  \midrule
      \multirow{4}{*}{Enc-Dec} & T5 3B & 82.45 & 84.80 & 50.30 & 0.57 & 0.08 & 43.64\\
      & T5 11B & 78.70 & 79.06 & 53.41 & 24.50 & 0.06 & 47.15\\
      & FLAN-T5 3B & 74.67 & 84.84 & 58.02 & 19.33 & 0.00 & 47.37\\
      & FLAN-T5 11B & 74.87 & 79.00 & 50.70 & 10.97 & 0.00 & 43.11\\
  \midrule
      \multirow{4}{*}{Dec-only} & LLaMA 7B & 46.77 & 55.54 & 29.55 & 2.95 & 0.00 & 26.96\\
      & LLaMA 13B & 38.07 & 59.88 & 32.51 & 16.85 & 0.00 & 29.46\\
      & BLOOM 3B & 20.76 & 19.65 & 10.82 & 14.53 & 0.00 & 13.15\\
      & BLOOM 7.1B & 20.90 & 34.78 & 20.15 & 15.00 & 0.00 &18.17\\
      & ChatGPT* & 64.25 & 71.17 & 34.40 & 55.33 & 46.04 & 54.24\\
  \bottomrule
  \end{tabular}
  \caption{Comparison of Large Language Models' Performance in Generalizing to Unseen Information Types and Task Forms. We include the average F1 scores for each model, computed across all tasks. *: ChatGPT was tested using direct in-context learning and was not trained on our dataset.
  }
  \label{tab:main}
\end{table*}

\paragraph{Analysis of Generalization to Unseen Information Types.}
In the generalization to unseen information types, Table~\ref{tab:main} demonstrates that models with an encoder-decoder architecture tend to outperform those with a decoder-only structure. Specifically, the T5 models with 3B and 11B parameters achieved F1 scores of 82.45 and 78.70 respectively in the entity extraction task. These scores significantly surpass the highest F1 score (64.25) achieved by ChatGPT in the decoder-only category. For trigger and argument extraction, T5 and FLAN-T5 models consistently perform well, with FLAN-T5 3B achieving the highest F1 score (58.02) in argument extraction among all models.

It is noteworthy that ChatGPT, which utilizes in-context learning and wasn't trained on our dataset, demonstrates respectable performance, particularly in entity and trigger extraction. This suggests that pre-trained models with large-scale knowledge can exhibit reasonable generalization even without specific fine-tuning.

\paragraph{Analysis of Generalization to Unseen  Task Forms.}
As for the generalization to unseen task forms, the performance of most models substantially declines. Notably, ChatGPT attains significantly better results compared to others in this category. With an F1 score of 55.33 in entity extraction and 46.04 in the ASTE task, ChatGPT exhibits the ability to adapt more efficiently to unfamiliar task forms. On the contrary, encoder-decoder models, which performed well in generalization to unseen information types, struggle considerably, with the T5 11B model obtaining the highest F1 score among them in entity extraction (24.50), but almost negligible performance in the ASTE task.

\subsection{In-depth Discussion}
\paragraph{Effectiveness of Encoder-Decoder Models in Information Types.}
The encoder-decoder models, particularly T5 and FLAN-T5, display commendable proficiency in generalizing to unseen information types. This can be attributed to the ability of encoder-decoder models to effectively capture the relationships between inputs and outputs, which is crucial for IE tasks. Furthermore, the availability of an encoder component might contribute to better representation learning, which aids in generalization.

\paragraph{Limited Generalization to New Task Forms.}
Despite the superior performance in information type generalization, encoder-decoder models exhibit restricted generalization capabilities when subjected to unfamiliar task forms. This might be due to the high specialization of these models to the training task forms, which in turn hampers their ability to adapt to new structures. ChatGPT, however, with its in-context learning, appears more flexible and can reasonably adapt to new task forms. This highlights the importance of model adaptability and flexibility in real-world applications where task forms might not always be consistent.

\paragraph{Performance is Not Always Proportional to Scale.}
The results also indicate that an increase in the number of parameters does not always lead to a proportional improvement in performance. For example, the T5 3B model outperforms the T5 11B model in entity extraction within unseen information types. This suggests that model capacity, though important, is not the sole factor in determining performance. Other factors such as model architecture, training data diversity, and learning techniques play a crucial role.

\paragraph{Decoder-Only Architectures Struggle More in Information Types.}
Decoder-only models such as LLaMa and BLOOM tend to struggle more in generalization to unseen information types as compared to encoder-decoder models. This could be due to their lack of an encoder component, which is important for understanding complex input structures that are common in IE tasks. However, ChatGPT demonstrates that decoder-only models with extensive pre-training and in-context learning can still achieve reasonable performance. This indicates that training methodology and in-context adaptation can play a significant role in improving the generalization of decoder-only models.

\section{Further Analysis}

\begin{figure}[htbp] 
  \centering
  \includegraphics[width=1.0\linewidth]{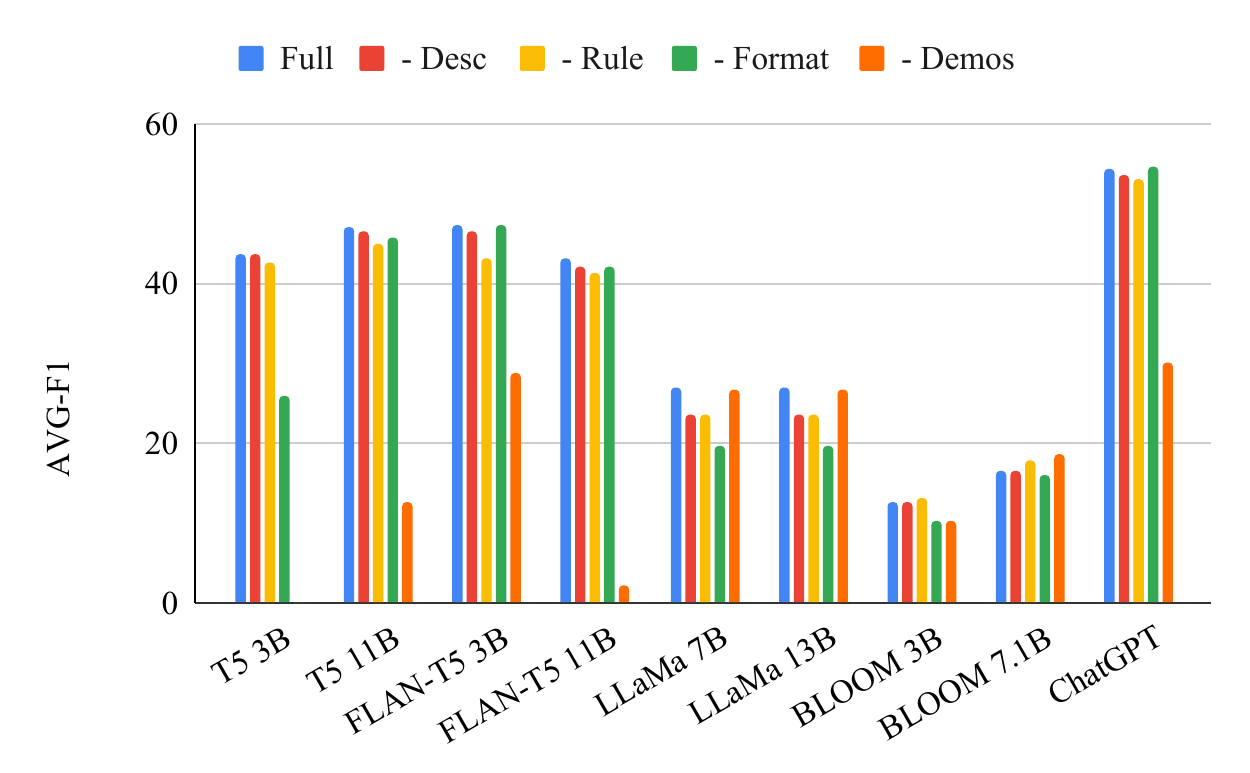}
  \caption{Impact of Instruction Components. The figure shows the average F1 scores of models with varying instruction components: Full (all components), -Desc (without Task Description), -Rule (without Extraction Rule), -Format (without Output Format), and -Demos (without Demonstration Examples).
  }
  \label{fig:exp_ablation_avg}
\end{figure}

\subsection{Impact of Instruction Components}
Figure~\ref{fig:exp_ablation_avg} presents the impact of various instruction components on the performance of LLMs in IE tasks. The components in consideration are \textsc{Task Description}, \textsc{Extraction Rule}, \textsc{Output Format}, and \textsc{Demonstration Examples}.

\paragraph{Task Description.}
The exclusion of the Task Description appears to have a marginal effect on the performance of the models. For example, T5 3B exhibits a slight increase from 43.64 to 43.79, and ChatGPT experiences a minor drop from 54.24 to 53.47. This suggests that while Task Description provides an overarching summary, it is not critical for performance. The Extraction Rule and Demonstration Examples likely offer the detailed guidance necessary for the models.

\paragraph{Extraction Rule.}
Omitting the Extraction Rule generally leads to a decrease in performance across most models. For instance, the T5 3B model drops from 43.64 to 42.49, and ChatGPT decreases from 54.24 to 52.96. This indicates that Extraction Rule, with its comprehensive guidelines, is crucial in guiding the models to extract the relevant information effectively.

\begin{figure}[htbp] 
  \centering
  \includegraphics[width=1.0\linewidth]{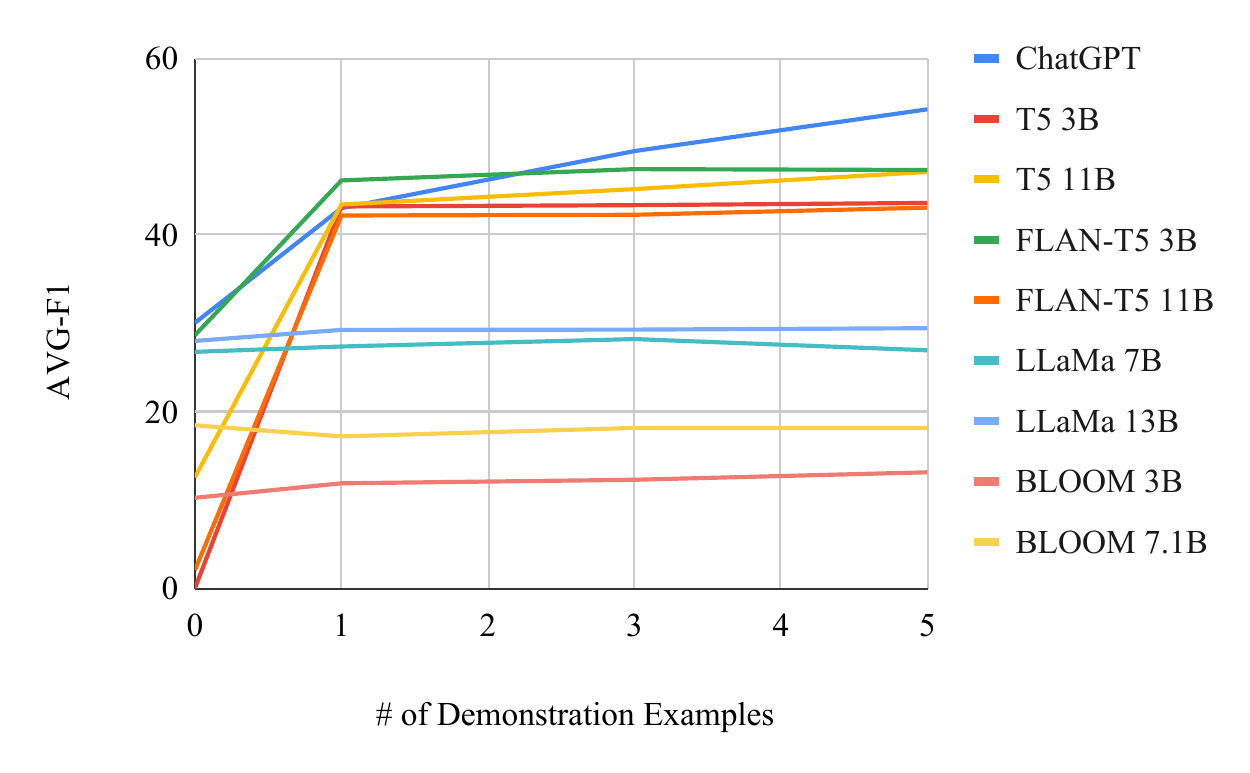}
  \caption{Impact of the number of examples.}
  \label{fig:exp_example_number}
\end{figure}

\paragraph{Output Format.}
The absence of Output Format leads to varied effects across the models. Notably, T5 3B shows a dramatic decrease from 43.64 to 25.84. On the other hand, ChatGPT experiences a minor increase in performance (54.67). This suggests that while Output Format is significant for structuring the output in some models, it may not be as crucial for others, especially if they have been pre-trained to handle diverse output structures.

\paragraph{Demonstration Examples.}
Removing Demonstration Examples has the most pronounced impact on performance. For example, T5 3B plummets from 43.64 to 0, and ChatGPT falls sharply from 54.24 to 30.06. This underscores the importance of Demonstration Examples in clarifying ambiguities and reinforcing the understanding of the task.

\subsection{Analysis on Demonstration Examples}

\paragraph{Impact of the number of examples.}
The impact of varying the number of demonstration examples on the performance of LLMs is shown in Figure~\ref{fig:exp_example_number}.
Across all models, it is evident that the provision of demonstration examples significantly influences performance, especially when transitioning from zero to one example. However, the effect of adding more examples varies among models. ChatGPT and T5 models show a consistent positive trend, while FLAN-T5, LLaMa, and BLOOM models exhibit varied patterns. This analysis highlights the importance of demonstration examples in IE tasks and suggests that the optimal number of examples can differ based on the model's architecture and capabilities.

\begin{figure}[htbp] 
  \centering
  \includegraphics[width=1.0\linewidth]{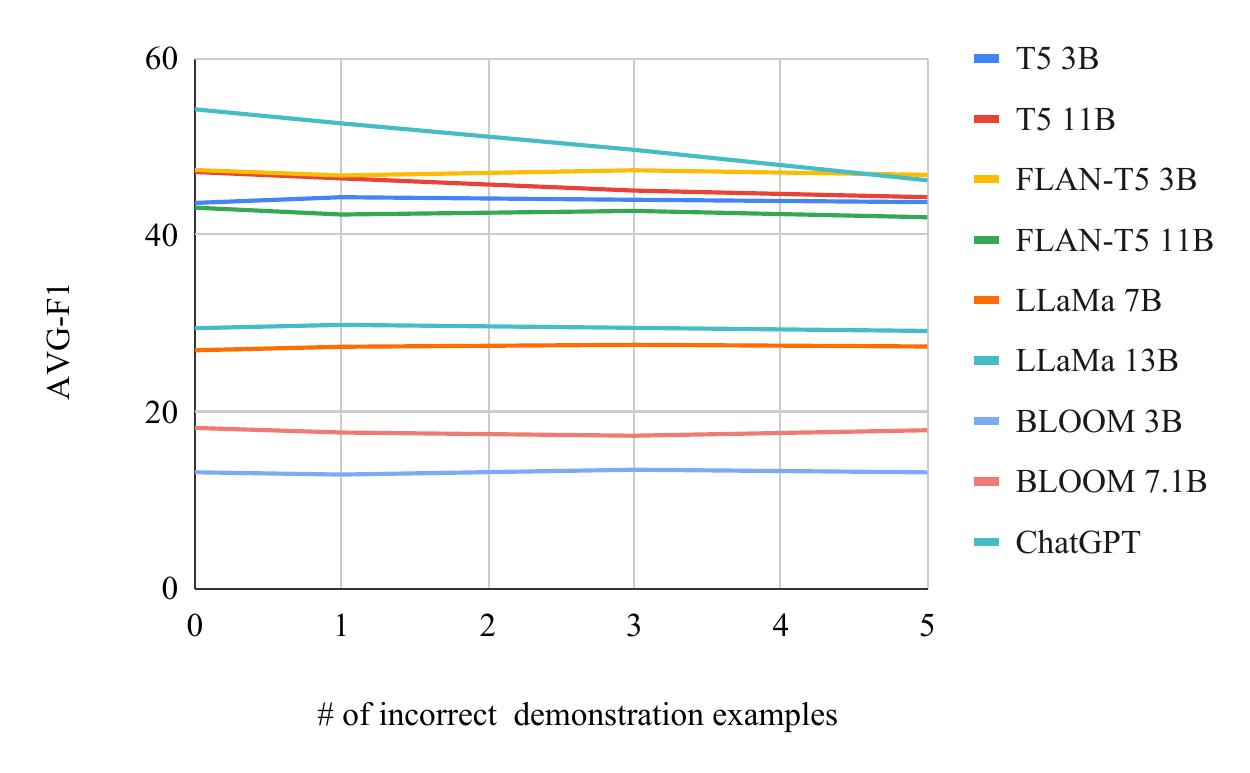}
  \caption{Impact of correctness of examples.}
  \label{fig:exp_example_quality}
\end{figure}

\paragraph{Impact of correctness of examples.}
The quality of demonstration examples is investigated by analyzing the performance of LLMs with different proportions of incorrect examples. Table~\ref{fig:exp_example_quality} presents the average F1 scores as we vary the number of incorrect demonstration examples. Across all models, the correctness of demonstration examples plays a vital role in performance. The sensitivity to incorrect examples, however, varies among models. ChatGPT is the most sensitive, with a pronounced decrease in performance as incorrect examples are introduced. T5 and FLAN-T5 models show stability or a gradual decline, while LLaMa and BLOOM models display minor fluctuations. This analysis underlines the importance of ensuring the accuracy and correctness of demonstration examples in the instructions provided to LLMs, especially for models like ChatGPT that exhibit high sensitivity to example quality.

\begin{figure}[htbp] 
  \centering
  \includegraphics[width=1.0\linewidth]{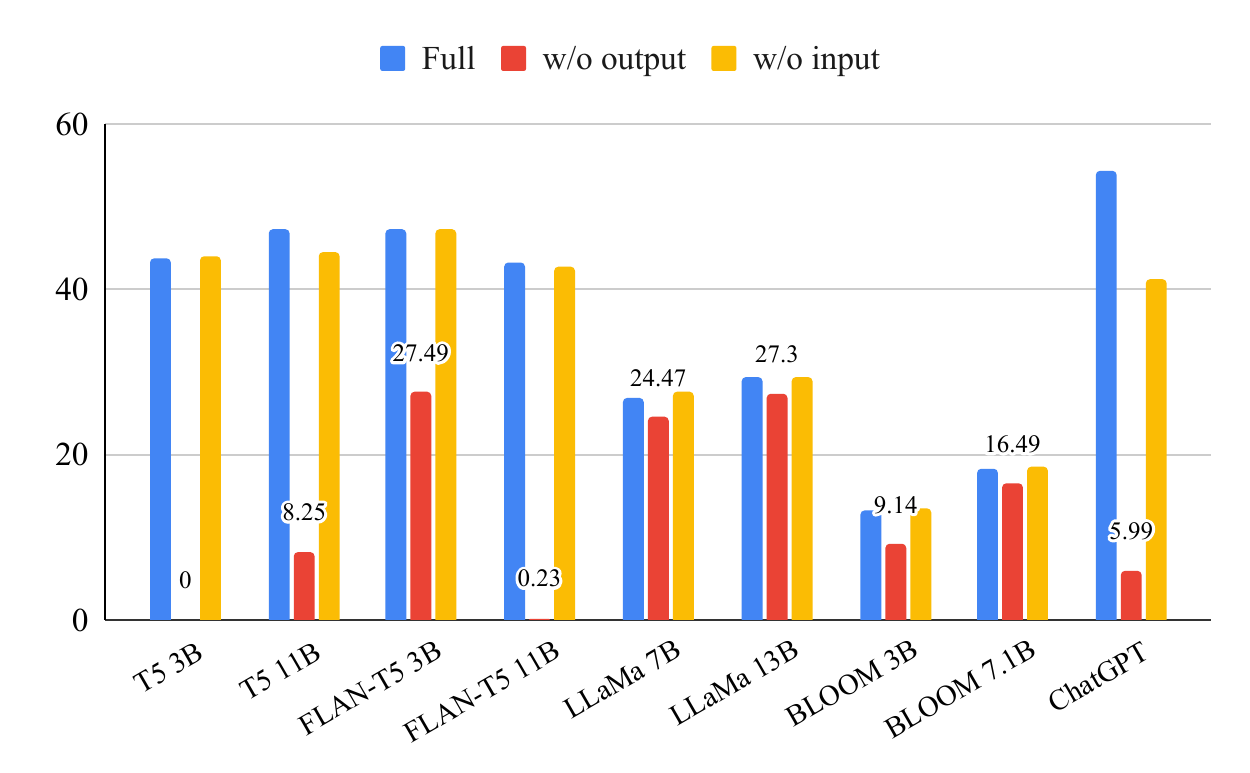}
  \caption{Impact of input-output pairing.}
  \label{fig:exp_example_pair}
\end{figure}

\paragraph{Impact of input-output pairing.}
Figure~\ref{fig:exp_example_pair} presents the performance of LLMs when they are conditioned on demonstration examples with varying formats - full examples with both inputs and outputs, examples without outputs, and examples without inputs. The goal is to understand which part of the demonstration example is crucial for performance.
Across all models, input-output pairing in demonstration examples plays a crucial role in performance. Models like FLAN-T5 11B and ChatGPT are heavily reliant on output information. LLaMa and BLOOM models also lean towards output information, whereas T5 models show variations. This analysis highlights the importance of including both inputs and outputs in demonstration examples for optimal performance. However, if one must be omitted, it appears that maintaining the outputs is generally more beneficial than retaining only the inputs. This may be due to the fact that the outputs often embody the essence of the task that the model needs to perform.

\subsection{Analysis of Scaling Factors}
\begin{figure}[h] 
  \centering
  \includegraphics[width=1.0\linewidth]{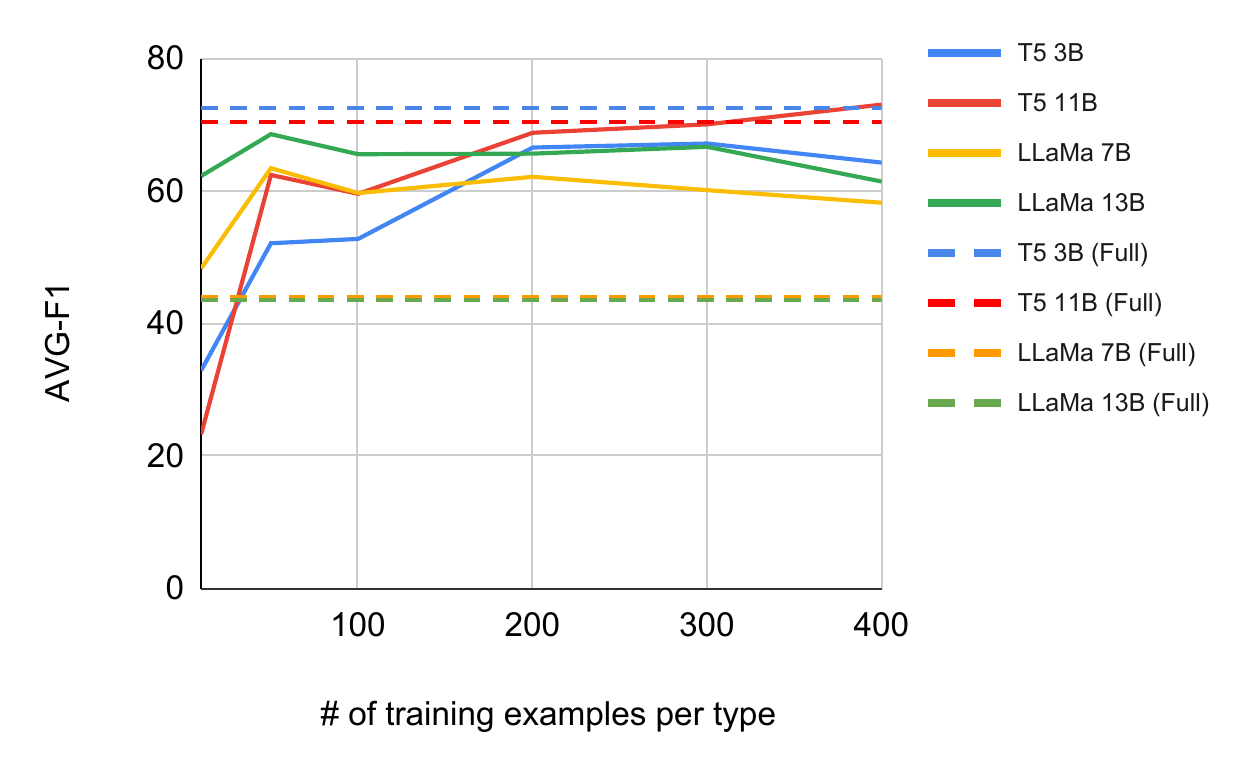}
  \caption{Impact of the number of training examples per information type.}
  \label{fig:exp_scale_trend}
\end{figure}

We analyze the generalization performance of the models with respect to two critical scaling factors: the number of instances per information type and the size of the models. The results are shown in Figure~\ref{fig:exp_scale_trend}. As the number of training instances is increased, note that not all information types are equally represented. Some event types have fewer than 100 samples. In such cases, the dataset uses the maximum number of available samples for those types. This leads to an exacerbation of the imbalance among different information types as the number of instances increases, which is an important consideration in the scaling trends.

\paragraph{Influence of Training Instance Quantity.}
Figure~\ref{fig:exp_scale_trend} reveals that augmenting the number of training instances is generally associated with improved performance. However, the models react differently to this scaling factor. T5 models display a more steady improvement as the number of instances increases, whereas LLaMa models experience an early peak followed by a decrease. The decline in LLaMa models' performance could be linked to the increasing imbalance in the dataset. As the dataset grows, the imbalance may cause models to become biased towards information types with more samples. Additionally, the non-linear scaling indicates that there may be a point of diminishing returns, after which additional data does not yield significant performance gains or may even be counterproductive.

\paragraph{Effect of Model Size on Performance.}
When analyzing the impact of model size, the results suggest that larger models typically have the edge. T5 models, for instance, exhibit more consistent improvements as their size increases. However, this comes with the caveat that larger models are more susceptible to overfitting, especially when the dataset is small or imbalanced. This risk is pertinent as data scarcity can make it difficult for larger models to effectively generalize. In the decoder-only category, the difference in performance between LLaMa 13B and LLaMa 7B is not pronounced at higher training sizes, highlighting that an increase in model size does not guarantee proportionate performance improvements. Consequently, a judicious balance between model size and the quantity and diversity of training data is essential to maximize generalization performance.

\section{Related Work}
\paragraph{Information Extraction.}
Previously, Information Extraction (IE) focused on task-specific models optimized for narrow objectives like Entity Extraction~\citep{Yan2021AUG, Wang2021ImprovingNE} and Event Extraction~\citep{Yan2021AUG, Wang2021ImprovingNE, Du2020EventEB, Lu2021Text2EventCS, gao2023mask}. However, their task-specific design inhibits knowledge sharing across various IE tasks \citep{Lin2020AJN}. This shortcoming paved the way for Universal Information Extraction (UIE), which aims at building versatile models for extracting diverse structured data from unstructured text \citep{Lin2020AJN, Lu2022UnifiedSG, Lou2023UniversalIE, Liu2023RexUIEAR}. Current UIE methods employ coarse-grained instructions with basic task descriptions, overlooking essential extraction rules and output format descriptions. To address this, we introduce a fine-grained benchmark dataset for IE with augmented instructions encompassing task descriptions, extraction rules, output formats, and demonstration examples.

\paragraph{Large Language Models.}
Large Language Models (LLMs) are central to NLP due to their impressive performance on numerous tasks \citep{devlin2018bert, radford2019language, lewis2019bart, raffel2020exploring, Brown2020LanguageMA, chowdhery2022palm}. Pretrained on vast corpora, LLMs can be fine-tuned for specialized tasks. Recently, instruction tuning has emerged, wherein LLMs are fine-tuned using task instructions for enhanced zero-shot task generalization \citep{sanh2021multitask, chung2022scaling, ouyang2022training}. By scaling training tasks, prompts, and LLM sizes, performance improves markedly. Combining instruction tuning with demonstration examples further optimizes results \citep{min2021metaicl, chen2021meta, ye2023guess}. In this work, we assess LLMs with different architectures (encoder-decoder and decoder-only) and ChatGPT for the IE task.

\section{Conclusion}
This paper introduced a fine-grained IE benchmark dataset, tailored for LLMs, utilizing augmented instructions to address the limitations of traditional coarse-grained IE. Through extensive evaluation, encoder-decoder models, notably T5 and FLAN-T5, showed prowess in generalizing to unseen information types, owing to their capacity for capturing complex input-output relationships. However, they exhibited limited adaptability to novel task forms. ChatGPT, a decoder-only model with in-context learning, demonstrated remarkable flexibility and adaptability.
Furthermore, we found that model scale is not the sole determinant of performance, emphasizing the importance of architecture, data diversity, and learning techniques.
our work contributes to the evolution of IE by enabling more refined IE through LLMs. Future endeavors should focus on combining the strengths of different architectures and devising training methods that optimize both specificity and adaptability in IE tasks.

\bibliography{custom}
\bibliographystyle{acl_natbib}

\appendix

\section{Implementation Details}
\label{sec:impl_detail}
\subsection{BLOOM and LLaMa Models}
We utilized the DeepSpeed library for the fine-tuning of both the LLaMa and BLOOM models, which include variants such as LLaMa-7B/13B and BLOOM-3/7.1B. DeepSpeed enabled us to perform distributed training with mixed precision support, effectively combining the benefits of memory efficiency and computational speed.
The training process employed a batch size of 8 and did not require gradient accumulation steps. For optimization, the AdamW optimizer was chosen, with a learning rate of 5e-5 and a weight decay parameter of 1e-4. To gradually reduce the learning rate during training, we employed a cosine annealing scheduler, while forgoing the use of any warm-up steps.

Additionally, the training made use of 16-bit floating-point precision (FP16), which is known to reduce memory usage while accelerating the training process.
DeepSpeed's Zero Redundancy Optimizer (ZeRO) was configured at stage 2, wherein the optimizer states were offloaded to the CPU memory, resulting in a further reduction in GPU memory consumption. To ensure consistency and reproducibility in training results, the random seed was fixed at 1024.
The training was carried out over two epochs, and model checkpoints were saved upon the completion of each epoch.

\subsection{T5 and FLAN-T5 Models}
For the fine-tuning of the T5 and FLAN-T5 models, including their variants T5-3B/11B and FLAN-T5-3B/11B, we similarly leveraged the DeepSpeed library to facilitate distributed training across eight GPUs.
In this case, the training was conducted with a batch size of 1 and gradient accumulation steps set to 8, effectively simulating a larger batch size while avoiding excessive memory consumption.
The optimization process mirrored that of the BLOOM and LLaMa models, using the AdamW optimizer with identical learning rate and weight decay parameters, and employing a cosine annealing learning rate scheduler without warm-up steps.

Both source and target sequence lengths were limited to 1024 tokens to maintain computational efficiency.
Interestingly, the training process was executed using the bfloat16 numerical format instead of FP16. This format is known to strike an optimal balance between training speed and numerical precision, thus proving advantageous in large-scale model training.
Like the BLOOM and LLaMa models, ZeRO stage 2 was applied to offload optimizer states to CPU memory, and a random seed of 1024 was utilized to assure the reproducibility of the training outcomes.

\section{Details of Benchmark Dataset}
\label{sec:data_collection}

\begin{figure*}[htbp] 
  \centering
  \includegraphics[width=1.0\linewidth]{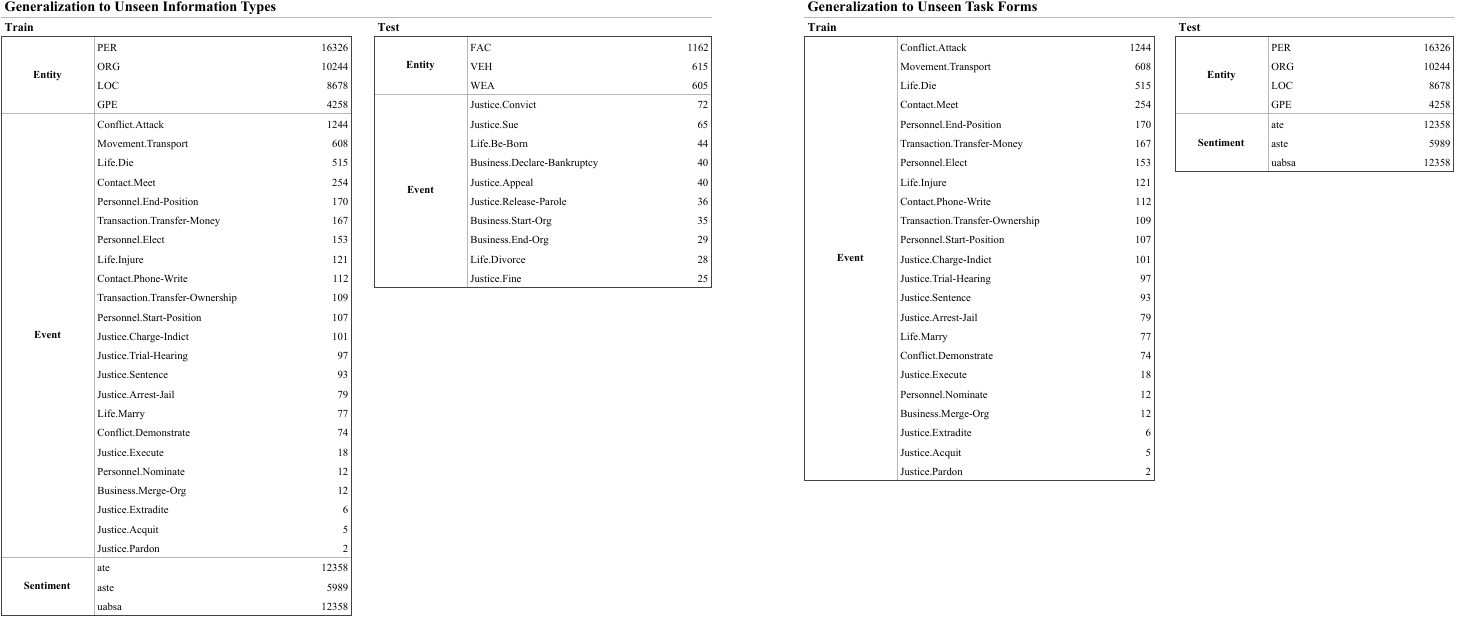}
  \caption{Detailed depiction of data partitioning across training and test sets in two distinct experimental configurations.}
  \label{fig:data_split_table}
\end{figure*}

\begin{figure*}[htbp] 
  \centering
  \includegraphics[width=1.0\linewidth]{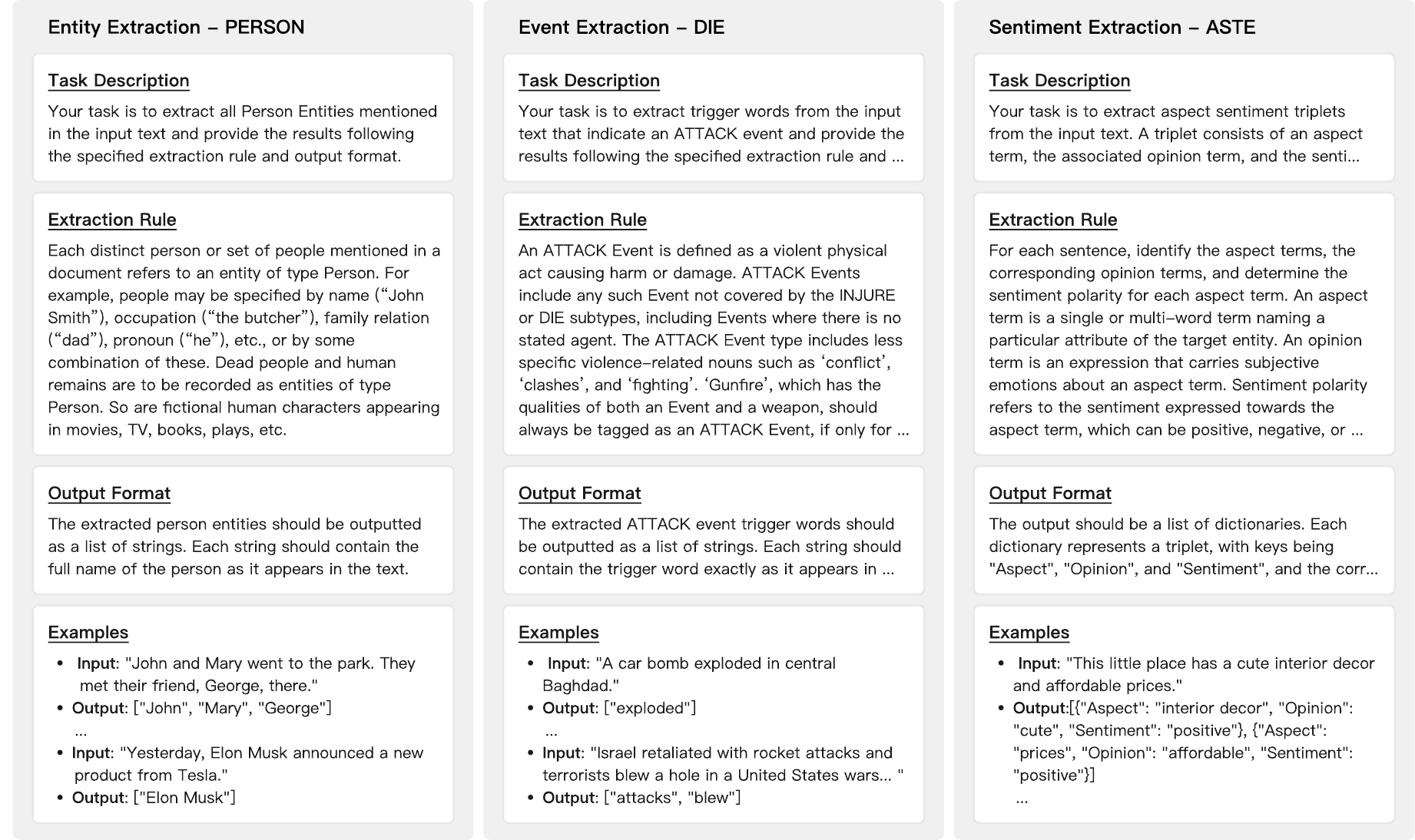}
  \caption{Detailed instructions for the three tasks. More instructions can be found via this link \url{https://anonymous.4open.science/r/IE-NAINST-6808}.}
  \label{fig:three_examples}
\end{figure*}

In our research, the data for events and entities is sourced from the ACE05 dataset, while the data for sentiment information extraction is derived from four datasets, namely 14lap, 14res, 15res, and 16res. To extract event and entity information with higher precision, we designed a set of extraction rules based on the event annotation guidelines of ACE05. Additionally, to ensure the accuracy of sentiment information extraction, we consulted an expert with extensive experience in the field of sentiment analysis to write the corresponding extraction rules.
To provide further insight into our data processing, Figure~\ref{fig:data_split_table} details the data partitioning in two different experimental settings. Figure~\ref{fig:three_examples}, on the other hand, showcases examples of the instructions that we wrote for three different tasks.



\end{document}